\documentclass[runningheads]{llncs}

\usepackage{eccv}

\usepackage{eccvabbrv}

\usepackage{graphicx}
\usepackage{booktabs}
\usepackage{multirow}
\usepackage{float}

\usepackage[accsupp]{axessibility}  

\usepackage{hyperref}

\usepackage{orcidlink}

\begin{document}

\title{SIGMA: Bridging Structural and Distributional Gaps for Vision Foundation Model Adaptation}
\titlerunning{SIGMA}

\author{Lingyu Xiong  \and
Jinjin Shi\thanks{Corresponding author} \and
Xuran Xu \and
Cong Luo \and
Runyu Shi \and
Ying Huang}

\authorrunning{Xiong. et al.}

\institute{Xiaomi Corporation}

\maketitle

\begin{abstract}

  Vision Foundation Models (VFMs) have demonstrated impressive representational capabilities. However, adapting them to downstream tasks via full fine-tuning incurs prohibitive computational and storage overhead. Parameter-Efficient Fine-Tuning (PEFT) has emerged as a compelling alternative, aiming to achieve performance parity with full fine-tuning at minimal training costs. Nonetheless, applying PEFT to VFMs for dense prediction tasks remains challenging due to the structural and distributional gaps. To bridge these gaps, we propose \textbf{S}cale-\textbf{I}ntegrated \textbf{G}lobal \textbf{M}odulation \textbf{A}dapter (\textbf{SIGMA}), a novel lightweight PEFT method, which consists of two modules: scale-adaptive fusion and semantic modulation. Specifically, the scale-adaptive fusion module is utilized to bridge structural gaps by enhancing the extraction of multi-granularity visual information. Furthermore, SIGMA introduces semantic modulation on the fusion features to perform global feature alignment to further eliminate the distribution gap. This design facilitates unified spatial and distributional adaptation, requiring only 1.72\% trainable parameters relative to the VFM backbone. Comprehensive experiments across various downstream dense tasks and multiple VFM backbones demonstrate that SIGMA achieves consistent and superior performance over state-of-the-art PEFT methods.

\end{abstract}

\section{Introduction}
\label{sec:intro}
The rapid evolution of Vision Transformers~\cite{dosovitskiy2020image, liu2021swin, fan2021multiscale, darcet2023vision} has fueled the rise of Vision Foundation Models (VFMs)~\cite{oquab2023dinov2, tschannen2025siglip, ravi2024sam, he2022masked}, which are trained on massive datasets with diverse pre-training objectives. Self-supervised approaches such as DINOv2~\cite{oquab2023dinov2} leverage self-distillation on curated data to produce robust, general-purpose visual features. Vision-language contrastive methods, notably by SigLIP2~\cite{tschannen2025siglip}, align visual and textual representations for open-vocabulary understanding. Meanwhile, Segment Anything Model (SAM)~\cite{ravi2024sam} establishes a promptable segmentation paradigm trained on billion-scale annotated data. These models encode rich visual knowledge that holds considerable promise for downstream dense prediction tasks, including object detection~\cite{tan2020efficientdet, redmon2016you, zhao2024detrs, huang2025deim}, semantic segmentation~\cite{yu2018bisenet, he2017mask, xiao2018unified, long2015fully}, and depth estimation~\cite{ranftl2021vision, yang2024depth_v2, ranftl2020towards}. However, directly deploying VFMs with frozen backbones suffers from inferior performance on these tasks~\cite{chen2025vfm}. While fully fine-tuning both the backbone and task-spefic head can improve performance, It demands prohibitive computational overhead and substantial training period. Moreover, maintaining a fully fine-tuned VFM for each downstream task is inherently inefficient and impractical in terms of storage and deployment. 

\begin{figure}[!t]
    \centering
    \includegraphics[width=1.0\linewidth]{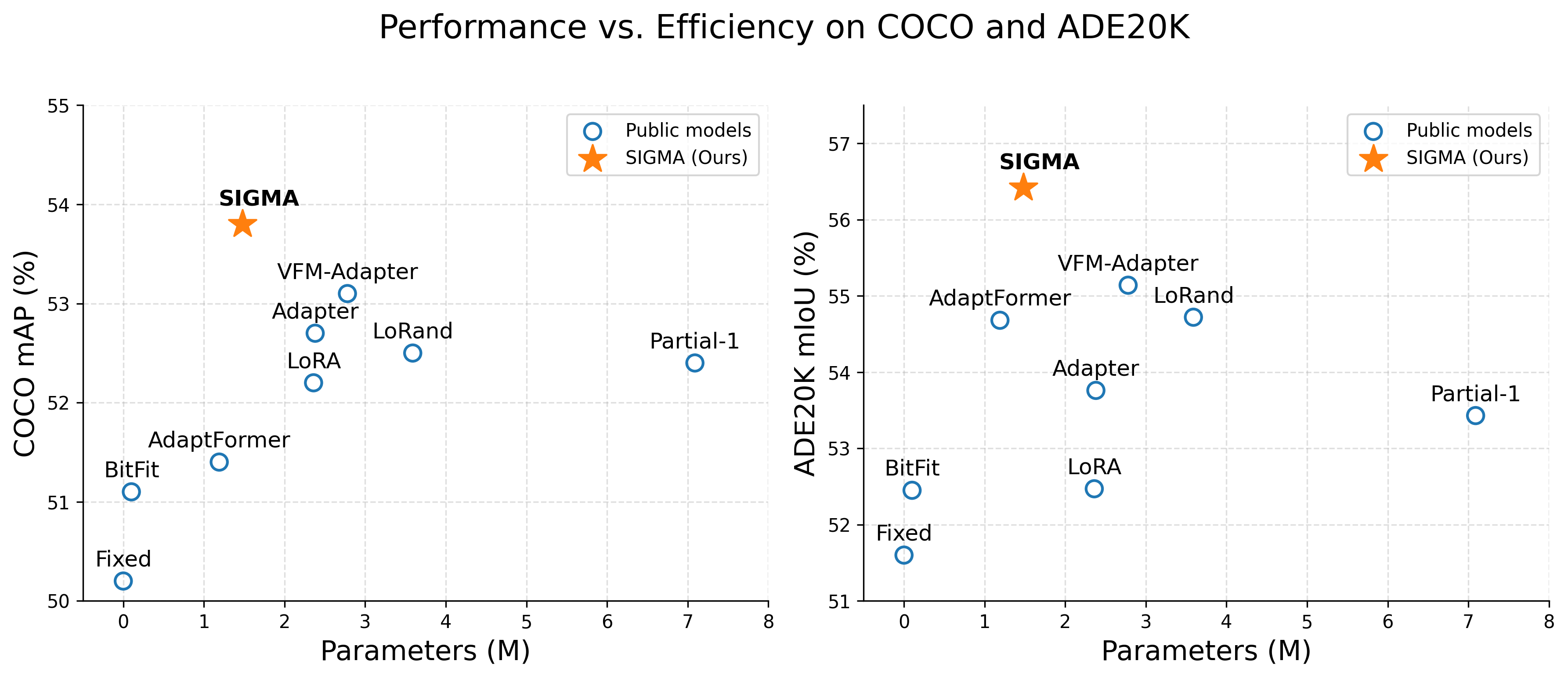}
    \caption{Comparison of our method with other tuning methods on representative visual tasks. SIGMA (orange star) significantly outperforms existing PEFT methods, achieving state-of-the-art performance with only 1.48M trainable parameters.}
    \label{fig:per_vs_effi}
\end{figure}

Parameter-Efficient Fine-Tuning (PEFT)~\cite{adapter, hu2022lora, lorand, zaken2022bitfit} has become the dominant paradigm for adapting foundation models, requiring only a small fraction of parameters to be updated while achieving performance comparable to full fine-tuning. These representative approaches include Adapter~\cite{adapter}, which inserts bottleneck layers into Transformer blocks; LoRA~\cite{hu2022lora}, which introduces low-rank matrices into attention layers. While these methods have achieved remarkable success in image classification and natural language processing, adapting VFMs for dense visual prediction remains challenging. We identify two gaps in directly applying existing PEFT methods to VFMs: a structural gap and a distributional gap. These gaps fundamentally limit their performance for dense prediction tasks.

The first is a structural gap: most existing PEFT methods originate from the natural language processing community, where adaptation operates on 1D token sequences through linear projections. Adapter and AdaptFormer~\cite{chen2022adaptformer} employ fully-connected bottleneck layers that process each token independently, while LoRA constrains weight updates to low-rank matrices that likewise lack spatial awareness. However, dense prediction tasks demand fine-grained spatial understanding. Without explicit mechanisms to capture 2D spatial locality and multi-scale context, these linear adapters are inherently ill-suited for dense visual tasks. Furthermore, the low-rank constraint in LoRA further limits the model's representational capacity for dense pixel-level predictions~\cite{chen2025vfm}.

The second is a distributional gap: VFMs are pre-trained on massive and diverse datasets (\eg, LVD-142M for DINOv2), and the resulting feature statistics, including the mean and variance of intermediate representations, inevitably diverge from those of specific downstream domains. Existing adapters lack explicit mechanisms to correct this statistical mismatch. Instead, they must implicitly learn distribution alignment within limited parameter budgets. While SSF~\cite{ssf} has shown that feature modulation can partially bridge this gap, applying such modulation alone proves insufficient for dense prediction, where features must be spatially and semantically aligned across multiple scales. 

To simultaneously bridge both gaps, we propose \textbf{SIGMA} (\textbf{S}cale-\textbf{I}ntegrated \textbf{G}lobal \textbf{M}odulation \textbf{A}dapter), a novel PEFT framework specifically designed for adapting VFMs to dense prediction tasks. SIGMA integrates two synergistic components within a unified architecture. First, a \textit{Scale-Adaptive Fusion} module employs multi-scale depth-wise convolutions followed by point-wise aggregation. This design incorporates local spatial inductive biases, enabling the adapter to capture multi-scale spatial features essential for dense prediction. Second, a \textit{Semantic Modulation} mechanism generates content-adaptive affine parameters that explicitly refine the feature statistics at each processing stage, aligning the pre-trained representations with the downstream distribution. Crucially, the two components are tightly coupled: the modulation operates on each convolution filter's output, enabling joint spatial-distributional adaptation at every scale. 

The entire module introduces only 1.48M trainable parameters, less than half of LoRand~\cite{lorand}, and is seamlessly inserted after both the multi-head attention and feed-forward network in each Transformer block. To validate the generality and superiority of our proposed method, we evaluate SIGMA on three representative VFMs across common dense prediction tasks, including object detection, semantic segmentation, and depth estimation. Extensive experiments demonstrate the superiority and generalization of our approach, as depicted in \cref{fig:per_vs_effi}. In summary, our contributions are summarized as follows:
\begin{itemize}
    \item We identify and analyze the structural and distributional gaps that limit the generalization of existing PEFT methods when adapting VFMs to dense prediction tasks, providing principled motivation for vision-specific adapter design.
    \item We propose SIGMA, a novel method that integrates scale-adaptive fusion and semantic modulation to jointly address both gaps with only 1.48M trainable parameters.
    \item Extensive experimental results demonstrate that SIGMA outperforms state-of-the-art methods on dense prediction tasks, including object detection, semantic segmentation, and depth estimation.
\end{itemize}

\section{Related Work}
\label{sec:related_work}

\subsection{Vision Foundation Models}
The evolution of vision backbones from Convolutional Neural Networks~\cite{he2016deep, krizhevsky2012imagenet} to Vision Transformers~\cite{dosovitskiy2020image, liu2021swin} has facilitated the training of foundation models with unprecedented scale and capability. Current Vision Foundation Models (VFMs) can be categorized by their pre-training objectives: self-supervised learning and vision-language pre-training. self-supervised approaches, such as BEiT~\cite{bao2021beit}, MAE~\cite{he2022masked}, and iBOT~\cite{zhou2021ibot}, reconstruct masked patches from visible ones, encouraging the model to learn context-aware representations. DINOv2~\cite{oquab2023dinov2} further advances self-supervised learning by training on the large-scale curated LVD-142M dataset, producing general-purpose features that perform well on visual tasks. In parallel, vision-language pre-training has evolved toward efficiency and scalability~\cite{li2023scaling, sun2023eva, vasu2024mobileclip}. While CLIP~\cite{radford2021learning} established the contrastive paradigm, SigLIP~\cite{zhai2023sigmoid} improves the efficiency of contrastive pre-training by replacing softmax normalization with a pairwise sigmoid loss. Its successor, SigLIP2~\cite{tschannen2025siglip}, further enhances this paradigm by incorporating dynamic resolution adaptation and self-distillation for improved open-vocabulary performance. Furthermore, Segment Anything Model (SAM)~\cite{kirillov2023segment, ravi2024sam} introduces a promptable segmentation framework trained on the billion-scale SA-1B dataset, enabling strong zero-shot generalization through point or box prompts. However, we empirically find that applying these models to dense prediction tasks such as object detection often suffer from inferior performance when the backbone is kept frozen. Consequently, parameter-efficient fine-tuning is essential for adapting these models to downstream tasks.

\subsection{Parameter-Efficient Fine-Tuning}
Parameter-Efficient Fine-Tuning (PEFT) aims to adapt pre-trained models to downstream tasks by updating only a small fraction of parameters~\cite{hu2022lora, adapter, lorand, zaken2022bitfit}. Existing approaches can be classified into three categories: parameter subset tuning, additive modules, and low-rank adaptations. Parameter subset methods tune specific components of the original backbone. BitFit~\cite{zaken2022bitfit} demonstrates that updating only the bias terms can achieve competitive performance. Similarly, Partial-1~\cite{partial-1} fine-tunes only the last Transformer block, serving as a strong baseline for transfer learning. Additive methods introduce extra learnable parameters into the backbone. The canonical Adapter~\cite{adapter} inserts bottleneck layers sequentially within Transformer blocks. Tailored for vision tasks, AdaptFormer~\cite{chen2022adaptformer} employs a parallel bottleneck structure to better preserve visual features. Inspired by NLP prompting, Visual Prompt Tuning (VPT)~\cite{vpt} prepends learnable tokens to the input sequence to guide the frozen backbone. Finally, low-rank adaptation methods assume that weight updates lie in a low-dimensional subspace. LoRA~\cite{hu2022lora} introduces trainable low-rank matrices into attention layers while keeping the pre-trained weights frozen. LoRand~\cite{lorand} employs a shared low-rank parameter decomposition strategy specifically optimized for dense prediction tasks, enabling competitive performance in detection and segmentation. ARC~\cite{arc} introduces adaptive rank control to dynamically allocate parameter budgets across layers based on sensitivity. Despite this diversity, most existing PEFT methods remain rooted in NLP design principles, relying on linear transformations that process tokens as 1D sequences. This renders them structurally agnostic to the 2D spatial locality and multi-scale context that are critical for dense prediction. Unlike existing methods, our work introduces scale-adaptive fusion and feature modulation modules into the tuning architecture, explicitly bridging the structural and distributional gaps.

\begin{figure}[!t]
    \centering
    \includegraphics[width=0.95\linewidth]{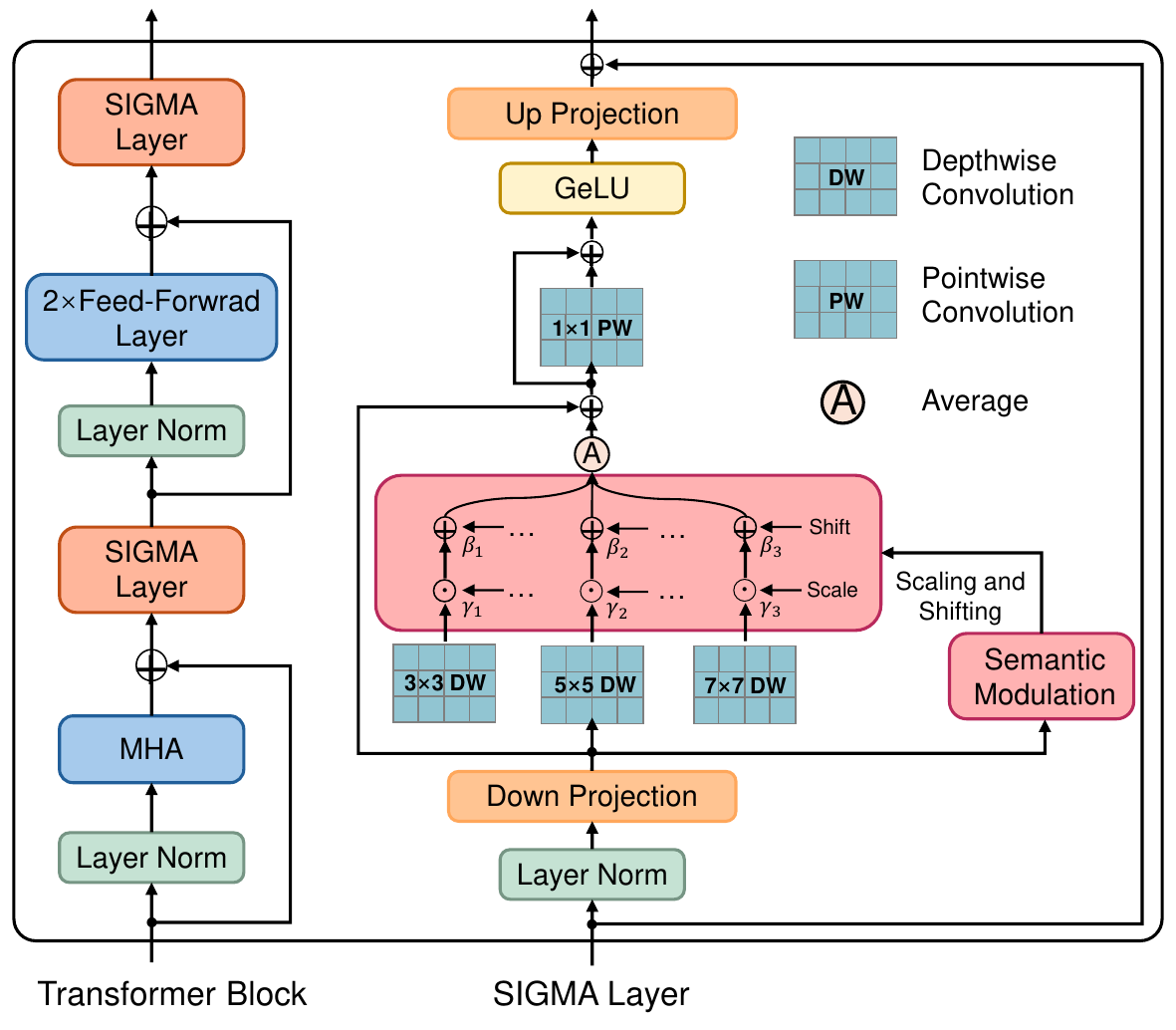}
    \caption{Overview of the proposed SIGMA tuning framework. \textbf{Left:} We add SIGMA layer after MHA and FFN in each Transformer block. The proposed method freezes the parameters of the original pre-trained layers and only updates the parameters within SIGMA layer. \textbf{Right:} The details of SIGMA layer. The process begins with LayerNorm and down-projection. Multi-scale and aggregation filters then sequentially process the features, while a Semantic Modulation module concurrently generates scale and shift signals to modulate these filters. The output is then processed with an activation function and up-projection to the original space. Four residual connections are included to improve adaptability.}
    \label{fig:framework}
\end{figure}

\subsection{Feature Modulation}
Modulating intermediate features is a canonical strategy in deep learning to regulate representation distributions. The most fundamental form lies in normalization layers, such as Batch Normalization (BN)~\cite{ioffe2015batch}, Layer Normalization (LN)~\cite{ba2016layer}, and Group Normalization (GN)~\cite{wu2018group}. These methods typically standardize features and subsequently apply learnable affine transformations to enhance training stability and generalization. Beyond stabilization, feature modulation has been extensively explored for conditional control and information fusion. In the realm of image generation, AdaIN~\cite{adain} and Self-Modulation~\cite{chen2018self} dynamically generate affine parameters from latent codes to inject style information into the synthesis process. Similarly, STN~\cite{stn} introduces a learnable module to explicitly modulate the spatial geometry of feature maps. In vision-language scenarios, mechanisms like Conditional BN~\cite{de2017modulating} and FiLM~\cite{perez2018film} are widely adopted to modulate visual features based on linguistic inputs, effectively fusing multi-modal information. More recently, this paradigm has been adapted for efficient fine-tuning. Unlike standard normalization, which aims for convergence within a single task, SSF~\cite{ssf} leverages linear modulation to bridge the distribution gap between pre-training and downstream tasks. By inserting lightweight scaling factors after existing operations, it achieves parameter efficiency without being restricted to normalization layers. However, SSF and related approaches apply modulation globally without accounting for the multi-scale spatial structure in dense prediction tasks. In contrast, SIGMA integrates feature modulation with scale-adaptive fusion, generating scale-adaptive scale and shift parameters that control the distribution of output features. This design enables simultaneous spatial and distributional adaptation within a single module.

\section{Method}
\label{sec:method}
In this section, we present our proposed method in detail. We first revisit the differences between full fine-tuning and adapter tuning in \cref{sec:prelim}. Subsequently, we provide a comprehensive description of our methodology, explaining the key components and mechanisms in \cref{sec:sm-adapter}. Finally, we conduct a parameter analysis to demonstrate the efficiency of our approach in \cref{sec:complexity}.

\subsection{Preliminary and Notation}
\label{sec:prelim}
Given a pre-trained Vision Transformer (or Backbone) parameterized by $\Theta$, let $x \in \mathbb{R}^{N\times d}$ denote the input features of a specific layer, where $N$ is the sequence length and $d$ is the hidden dimension.

\textbf{Full fine-tuning} serves as the standard paradigm for adapting pre-trained models to downstream tasks. In this setting, we initialize the model with the pre-trained weights $\Theta_{pre}$ and update all parameters during training. Formally, the optimization objective is:
\begin{equation}
    \Theta^* = \arg \min_\Theta \sum_{(x,y)\in \mathcal{D}} \mathcal{L}(f(x;\Theta), y),
\end{equation}
where $\mathcal{D}$ represents the downstream dataset and $\mathcal{L}$ is the loss function. While full fine-tuning typically achieves superior performance, it is computationally expensive and parameter-inefficient, as the dimension of gradients $\Delta_\Theta \mathcal{L}$ scales with the entire model size.

\textbf{Adapter Tuning} freezes the pre-trained backbone and introduces a small set of learnable parameters. Formally, we define the model parameters as a tuple $(\Theta, \mathcal{W})$, where $\Theta$ represents the pre-trained weights and $\mathcal{W}$ denotes the newly introduced adapter parameters. Unlike full fine-tuning, the pre-trained parameters are kept frozen, and the training objective is reformulated as:
\begin{equation}
    \mathcal{W}^* = \arg \min_\mathcal{W} \sum_{(x,y)\in \mathcal{D}} \mathcal{L}(f(x;\Theta, \mathcal{W}), y).
\end{equation}

\subsection{SIGMA}
\label{sec:sm-adapter}
Although typical Linear Adapter and LoRA have shown great promise, they face specific challenges when applied to dense prediction tasks. First, VFMs are typically pre-trained on massive general datasets. It is difficult to effectively transfer these pre-trained representations to the specific distributions of downstream tasks using only simple linear layers. Second, the original Adapter and LoRA were designed for NLP and lack the structural inductive biases to process spatial visual information effectively. Third, prior work such as VFM-Adapter~\cite{chen2025vfm} suggests that the low-rank assumption in LoRA restricts the model's capacity for dense pixel-level predictions. To address these issues, we propose two key strategies, which are illustrated in \cref{fig:framework}. First, we introduce local inductive biases specifically tailored for dense prediction, which will be detailed in the Scale-Adaptive Fusion section. Second, to bridge the distribution gap, we employ learnable scale and shift operations to modulate the output features, aligning them with the downstream data distribution. 

\subsubsection{Scale-Adaptive Fusion.} To achieve superior performance in dense prediction tasks, it requires model to handle objects of varying scales while capturing spatial context across different regions. Therefore, we introduce scale-adaptive filters designed to extract and aggregate multi-scale semantic features. Instead of standard convolutions, which incur high computational costs, we employ Depth-wise (DW) convolutions to process the features efficiently. Specifically, let $x_0$ denote the original input and $x_r$ denote the input features after down-projection. We employ three parallel DW convolutions with kernel sizes of $3\times3$, $5\times5$, and $7\times7$ to extract spatial features at different receptive fields. These multi-scale features are then aggregated through a $1\times1$ Point-wise (PW) convolution. To preserve the feature identity and facilitate gradient flow, a residual connection is added between the DW and PW layers. The above steps can be formulated as:
\begin{equation}
    x_r = \text{Down}(|x_0|_{LN}),
\end{equation}

\begin{equation}
    x_{ms} = x_r + \text{avg} \Big( \sum_{k\in { \{ 3,5,7 \} } } \text{DWConv}_k(x_r) \Big),
\end{equation}
\begin{equation}
    x_{agg} = x_{ms} + \text{PWConv}(x_{ms}),
\end{equation}

where $|\cdot|_{LN}$ denotes LayerNorm, $\text{Down}(\cdot)$ denotes down-projection, $\text{DWConv}(\cdot)$, $\text{PWConv}(\cdot)$ denote depth-wise and point-wise convolution, respectively. Finally, the fused features pass through a GeLU activation function and an up-projection to restore the original dimension $d$. The overall computation process is formulated as:
\begin{equation}
    x = x_0 + \text{Up}(\text{GeLU}(x_{agg})),
\end{equation}
where $x_{agg}$ denotes the aggregated features after PW convolution, $\text{Up}(\cdot)$ denotes up-projection, and $\text{GeLU}(\cdot)$ denotes GeLU activation.

\subsubsection{Semantic Modulation.} Another fundamental challenge in adapting VFMs is the distribution discrepancy between the pre-training datasets and the downstream tasks. The feature statistics learned during pre-training often misalign with the downstream data distribution. Inspired by feature modulation techniques, we employ a scaling and shifting modulation mechanism to bridge this gap. We introduce learnable scale $\gamma$ and shift $\beta$ parameters that modulate the extracted features toward the downstream distribution. As illustrated in \cref{fig:framework}, the Semantic Modulation module operates immediately after the down-projection layer, dynamically generating four groups of affine parameters. These parameters modulate the outputs of the three DW convolutions and the final $1\times1$ PW convolution. Let $x_r$ denote the down-projected input. The modulation parameters are generated through a linear projection $\mathcal{P}$:
\begin{equation}
    [ \gamma_k, \beta_k]_{k \in {\{ 3,5,7,pw \}}} = \mathcal{P}(x_r),
\end{equation}
where $\gamma_k$ and $\beta_k$ denote the scale and shift parameters for $k$-th convolution. We define the modulation operation for a feature map $F$ as $\text{Mod}(F,\gamma, \beta)=\gamma \odot F + \beta$. The complete modulation and fusion process is formulated as:
\begin{equation}
    \hat{x}_{dw} = \text{avg} \Big( \sum_{k\in { \{ 3,5,7 \} }} \text{Mod}(\text{DWConv}_k(x_r), \gamma_k, \beta_k) \Big),
\end{equation}
\begin{equation}
    \hat{x}_{pw} = \text{Mod}(\text{PWConv}(\hat{x}_{dw}), \gamma_{pw}, \beta_{pw}),
\end{equation}

By explicitly adjusting the feature statistics at multiple stages, SIGMA effectively aligns the pre-trained representation space with the downstream task distribution.

\subsection{Complexity Analysis}
\label{sec:complexity}
In this section, we analyze the parameter efficiency of our proposed method. The parameters of SIGMA are primarily derived from the Layer Normalization (LN), Down and Up projection layers, Depth-wise (DW) convolutions, Point-wise(PW) convolutions, and the Semantic Modulation layer. Let the input feature be denoted as $x \in \mathbb{R}^{N \times d}$, where $N$ is the sequence length and $d$ is the hidden dimension. We set the bottleneck dimension as $r$(where $r \ll d$). The down-projection and up-projection layers contribute $2dr+d+r$ parameters. The LN introduces $2d$ learnable parameters. We employ three parallel DW convolutions with kernel sizes of $3\times3$, $5\times5$, and $7\times7$ operating in the bottleneck dimension. This contributes $(3^2+5^2+7^2)=83r$ parameters. An internal PW convolution is used for feature aggregation, adding $r^2$ parameters. The modulation module generates four groups of modulation weights (scale and shift). With a projection mapping, this contributes approximately $8r^2$ parameters. The total parameters of each SIGMA module are:
\begin{equation}
    9r^2 + (2d+84)r + 3d
\end{equation}
To maximize parameter efficiency, we set the bottleneck dimension $r=32$. Since $r \ll d$ (e.g., $d=768$ in ViT), the quadratic term $r^2$ remains small, and the complexity is dominated by the linear term $dr$, ensuring the adapter remains lightweight.

\section{Experiments}
\label{sec:exp}
We conduct sufficient experiments on multiple representative dense visual tasks to demonstrate the superiority of SIGMA. This section includes experimental setup, main results, and ablation studies.

\subsection{Experimental Setup}
\subsubsection{Dataset.} We evaluate our method on three downstream dense prediction tasks using standard benchmarks and frameworks. For object detection, we utilize the MS-COCO 2017 dataset~\cite{coco}. We adopt the DETR framework~\cite{detr, plain-detr}, training on the train2017 split and reporting mean Average Precision (mAP) on the val2017 split. For semantic segmentation, experiments are conducted on the ADE20K dataset~\cite{ade20k} with 150 categories. We employ Mask2Former~\cite{mask2former} implemented in MMSegmentation~\cite{mmseg2020} as the segmentation head, reporting mean Intersection over Union (mIoU) on the validation set. For depth estimation, we use the NYUv2 dataset~\cite{nyuv2}. Based on the DPT framework~\cite{dpt, depth_anything}, we evaluate the model using Root Mean Squared Error (RMSE) on the standard test split.

\subsubsection{Pretrained Backbone.} Our experiments are conducted on three distinct vision transformer backbones to ensure broad applicability, including DINOv2, SigLIP2, and SAM. Unless otherwise stated, we use the ViT-Base variant for all backbones to ensure a fair comparison of parameter efficiency.

\subsubsection{Baselines.} We benchmark our proposed approach against recent state-of-the-art methods.
\begin{itemize}
    \item Full tuning: Updates all parameters of the backbone and the head end-to-end.
    \item Fixed: Freezes the pre-trained backbone and only updates the task-specific head.
    \item BitFit~\cite{zaken2022bitfit}: Fine-tunes only the bias terms of the backbone while keeping the rest frozen.
    \item Partial-1~\cite{partial-1}: Updates only the last transformer block of the backbone.
    \item LoRA~\cite{hu2022lora}: Injects trainable low-rank matrices into the attention layers to approximate weight updates.
    \item Adapter~\cite{adapter}: Inserts sequential bottleneck modules into the transformer layers.
    \item AdaptFormer~\cite{chen2022adaptformer}: Adopts a parallel adapter architecture with a scaling factor in each MLP layer.
    \item LoRand~\cite{lorand}: Inserts LoRand layers after MHA/FFN layers in each transformer block.
    \item VFM-Adapter~\cite{chen2025vfm}: Leverages a HyperNetwork to dynamically synthesize convolutional parameters of the adapter for each transformer block.
\end{itemize}

\subsubsection{Implementation Details.} Unlike hierarchical models, DINOv2 and SigLIP2 are based on the plain, non-hierarchical Vision Transformer architecture, which produces single-scale feature maps. However, dense prediction heads (\eg, DETR and Mask2Former) typically require multi-scale inputs. To address this, we construct a multi-scale feature pyramid based on the feature maps from the last layer of the backbone. These features are upsampled or downsampled through convolutions to serve as inputs for the task-specific decoders. 

For COCO dataset, Input images are processed using random cropping and resizing to a unified scale of $640 \times 640$. We use the AdamW optimizer with an initial learning rate of $1\times 10^{-4}$ and weight decay of 0.05. The model is trained for 30 epochs with a linear warmup of 1000 steps. The learning rate is decayed by a factor of 10 at the 25th epoch. The total batch size is set to 32. For ADE20k, we follow the standard Mask2Former configuration provided by the MMSegmentation toolkit. The models are trained for 160k iterations with a total batch size of 16. For NYUv2, Input images are resized to a resolution of 640×480. We optimize the model using AdamW with a learning rate of $1.5 \times 10^{-4}$ and weight decay of 0.01. The training spans 20 epochs with a total batch size of 16. Experiments for object detection and semantic segmentation are conducted on 8 × NVIDIA Tesla A800 GPUs, while depth estimation tasks are performed on a single NVIDIA Tesla A800 GPU. Unless otherwise specified, the task-specific heads (\eg, the DETR decoder) are trained from scratch while the backbone is tuned according to each respective setting.

\subsection{Main Results}

\begin{table}[!htb]
    \caption{Quantitative comparison on the COCO object detection benchmark across DINOv2, SigLIP2, and SAM backbones. SIGMA consistently achieves the highest mAP among parameter-efficient methods while using significantly fewer trainable parameters than other competitive baselines like LoRand and VFM-Adapter.}
    \label{tab:quan_detection}
    \centering
    \begin{tabular}{l|c|c|ccccc}
    \toprule
    \multirow{2}{*}{\textbf{Backbone}} & \multirow{2}{*}{\textbf{Method}} & \textbf{Trained} & \multirow{2}{*}{mAP} & \multirow{2}{*}{$\text{mAP}_{50}$} & \multirow{2}{*}{$\text{AP}_s$} & \multirow{2}{*}{$\text{AP}_m$} & \multirow{2}{*}{$\text{AP}_l$} \\
    & & \textbf{Params} \\ \midrule
     \multirow{10}{*}{DINOv2}& Full tuning & 85.70M & 55.7 & 75.8 & 34.4 & 60.5 & 75.8 \\ \cmidrule{2-8} 
     & Fixed & 0.00M & 50.6 & 71.8 & 29.8 & 54.8 & 70.8 \\ 
     & BitFit & 0.10M & 51.2 & 71.8 & 30.5 & 55.4 & 71.6 \\ 
     & Partial-1 & 7.09M & 52.6 & 73.4 & 31.8 & 57.1 & 72.6 \\ 
     & LoRA & 2.36M & 52.2 & 72.7 & 31.5 & 56.6 & 72.5 \\ 
     & Adapter & 2.38M & 52.7 & 73.5 & 32.1 & 57.2 & 72.9 \\ 
     & AdaptFormer & 1.19M & 51.8 & 72.6 & 31.2 & 56.1 & 72.2 \\  
     & LoRand & 3.59M & 52.7 & 73.4 & 32.1 & 57.3 & 73.1 \\ 
     & VFM-Adapter & 2.78M & 53.2 & 73.8 & 32.6 & 57.7 & 73.6 \\ \cmidrule{2-8} 
     & SIGMA & 1.48M & \textbf{54.4} & \textbf{74.6} & \textbf{33.5} & \textbf{59.2} & \textbf{74.3} \\ \midrule
     \multirow{10}{*}{SigLIP2} & Full tuning & 86.43M & 54.6 & 74.8 & 33.6 & 59.2 & 74.6 \\ \cmidrule{2-8} 
     & Fixed & 0.00M & 50.2 & 71.2 & 28.6 & 54.4 & 70.2 \\  
     & BitFit & 0.10M & 51.1 & 72.1 & 29.4 & 55.2 & 71.0 \\ 
     & Partial-1 & 7.09M & 52.4 & 73.1 & 30.6 & 56.8 & 72.4 \\ 
     & LoRA & 2.36M & 52.2 & 72.9 & 30.8 & 56.7 & 72.1 \\ 
     & Adapter & 2.38M & 52.7 & 73.4 & 31.2 & 57.3 & 72.7 \\ 
     & AdaptFormer & 1.19M & 51.4 & 72.5 & 30.2 & 56.4 & 71.3 \\ 
     & LoRand & 3.59M & 52.5 & 73.1 & 30.9 & 57.2 & 72.5 \\
     & VFM-Adapter & 2.78M & 53.1 & 73.6 & 31.5 & 57.8 & 73.0 \\ \cmidrule{2-8} 
     & SIGMA & 1.48M & \textbf{53.8} & \textbf{74.2} & \textbf{32.1} & \textbf{58.7} & \textbf{73.7} \\ \midrule
     \multirow{10}{*}{SAM} & Full tuning & 85.80M & 52.6 & 73.6 & 31.8 & 56.8 & 72.8 \\ \cmidrule{2-8} 
     & Fixed & 0.00M & 48.3 & 69.2 & 27.8 & 52.1 & 69.4 \\ 
     & BitFit & 0.10M & 48.8 & 69.4 & 28.3 & 52.7 & 69.8 \\ 
     & Partial-1 & 7.09M & 49.5 & 70.5 & 29.1 & 53.4 & 70.7 \\ 
     & LoRA & 2.36M & 49.4 & 70.4 & 28.8 & 53.3 & 70.6 \\ 
     & Adapter & 2.38M & 49.3 & 70.2 & 28.6 & 53.2 & 70.6 \\ 
     & AdaptFormer & 1.19M & 50.2 & 71.2 & 29.4 & 54.3 & 71.5 \\ 
     & LoRand & 3.59M & 49.6 & 70.6 & 29.0 & 53.8 & 70.8 \\ 
     & VFM-Adapter & 2.78M & 50.3 & 71.4 & 29.5 & 54.5 & 71.6 \\ \cmidrule{2-8} 
     & SIGMA & 1.48M & \textbf{51.4} & \textbf{72.6} & \textbf{30.7} & \textbf{55.4} & \textbf{72.5} \\ \bottomrule
    \end{tabular}
    \vspace{-1em}
\end{table}

\begin{table}[!htb]
    \caption{Quantitative results of semantic segmentation on the ADE20K dataset. We report three metrics: mIoU (mean Intersection over Union), mAcc (mean Accuracy), and aAcc (average Accuracy). SIGMA outperforms state-of-the-art PEFT baselines across multiple backbones.}
    \label{tab:quan_segm}
    \centering
    \begin{tabular}{l|ccc|ccc|ccc}
    \toprule
    \multirow{2}{*}{\textbf{Method}} & \multicolumn{3}{c|}{\textbf{DINOv2}} & \multicolumn{3}{c|}{\textbf{SigLIP2}} & \multicolumn{3}{c}{\textbf{SAM}} \\
     & mIoU & mAcc & aAcc & mIoU & mAcc & aAcc & mIoU & mAcc & aAcc \\
    \midrule
    Full tuning & 57.10 & 70.50 & 85.50 & 55.89 & 69.10 & 84.85 & 49.54 & 61.45 & 82.35 \\
    \midrule
    Fixed & 51.60 & 67.80 & 83.70 & 51.66 & 66.54 & 83.04 & 45.72 & 59.32 & 81.21 \\
    BitFit & 52.45 & 67.62 & 84.63 & 52.71 & 67.93 & 84.21 & 46.28 & 59.6 & 81.35 \\
    Partial-1 & 53.43 & 67.63 & 84.57 & 53.56 & 67.26 & 84.12 & 47.18 & 60.10 & 81.55 \\
    LoRA & 52.47 & 67.60 & 84.23 & 52.47 & 66.31 & 83.67 & 46.36 & 59.65 & 81.40 \\
    Adapter & 53.76 & 67.92 & 84.46 & 53.77 & 67.07 & 84.35 & 46.52 & 59.75 & 81.45 \\
    Adaptformer & 54.68 & 68.18 & 84.63 & 53.81 & 67.77 & 84.41 & 46.64 & 59.8 & 81.48 \\
    Lorand & 54.72 & 68.25 & 84.34 & 53.84 & 67.42 & 84.53 & 47.05 & 60.05 & 81.52  \\
    VFM-Adapter & 55.14 & 68.50 & 84.85 & 54.36 & 67.85 & 84.45 & 47.36 & 60.25 & 81.60 \\ \midrule
    Ours & \textbf{56.42} & \textbf{69.28} & \textbf{85.06} & \textbf{55.12} & \textbf{68.23} & \textbf{84.72} & \textbf{48.58} & \textbf{61.1} & \textbf{81.95} \\
    \bottomrule
    \end{tabular}
\end{table}

\paragraph{Superior Performance Across Diverse Vision Tasks.}
\cref{tab:quan_detection}, \cref{tab:quan_segm}, and \cref{tab:quan_depth} show the results of objection, semantic segmentation, and depth estimation, respectively. SIGMA achieves the best performance among all PEFT methods across all tasks and backbones. On COCO object detection with DINOv2, SIGMA achieves 54.4 mAP, outperforming VFM-Adapter (53.2) by 1.2 mAP, and both LoRand and Adapter (52.7) by 1.7 mAP. The gap widens further for methods relying solely on linear transformations: LoRA (52.2) and AdaptFormer (51.8) trail by 2.2 and 2.6 mAP, respectively. Parameter subset methods such as BitFit (51.2) and Partial-1 (52.6) also underperform, since tuning only bias terms or a single block provides limited capacity for spatial adaptation. Notably, even VFM-Adapter, which introduces convolutional parameters through a HyperNetwork, falls behind SIGMA by 1.2 mAP on DINOv2 and 1.1 mAP on SAM, indicating that dynamically generated convolutions without explicit distributional alignment remain insufficient for dense prediction. These trends hold consistently across other tasks. On ADE20K segmentation (\cref{tab:quan_segm}), SIGMA surpasses VFM-Adapter by 1.28, 0.76, and 1.22 mIoU on DINOv2, SigLIP2, and SAM, respectively. On NYUv2 depth estimation (\cref{tab:quan_depth}), SIGMA reduces RMSE by 0.022, 0.011, and 0.009 over the strongest baseline on each backbone.

\paragraph{Trade-Off in Performance and Efficiency.}
SIGMA achieves superior results with only 1.48M trainable parameters, substantially fewer than most competing methods. Compared to LoRand (3.59M), SIGMA uses 2.4$\times$ fewer parameters yet outperforms it by 1.7 mAP on detection, 1.7 mIoU on segmentation, and 0.022 RMSE on depth estimation. Relative to VFM-Adapter (2.78M), SIGMA reduces the parameter count by 47\% while consistently delivering higher accuracy across all tasks. Even Partial-1, which tunes 7.09M parameters (roughly 4.8$\times$ more than SIGMA), only reaches 52.6 mAP on DINOv2 detection compared to 54.4 for SIGMA. Such results underscore the effectiveness of our architectural design, confirming that the improvements stem from the synergistic integration of multi-scale spatial filters and distributional modulation instead of increased model capacity.

\begin{table}[!t]
    \caption{Quantitative comparison of depth estimation on the NYUv2 dataset. SIGMA achieves the highest accuracy ($\delta_1$) and lowest error (AbsRel, RMSE) among PEFT methods, effectively narrowing the gap with full fine-tuning.}
    \label{tab:quan_depth}
    \centering
    \begin{tabular}{l|ccc|ccc|ccc}
    \toprule
    \multirow{2}{*}{\textbf{Method}} & \multicolumn{3}{c|}{\textbf{DINOv2}} & \multicolumn{3}{c|}{\textbf{SigLIP2}} & \multicolumn{3}{c}{\textbf{SAM}} \\
     & $\delta_1 \uparrow$ & AbsRel$\downarrow$ & RMSE$\downarrow$ & $\delta_1 \uparrow$ & AbsRel$\downarrow$ & RMSE$\downarrow$ & $\delta_1 \uparrow$ & AbsRel$\downarrow$ & RMSE$\downarrow$ \\
    \midrule
    Full tuning & 0.968 & 0.068 & 0.245 & 0.947 & 0.078 & 0.294 & 0.925 & 0.088 & 0.321 \\
    \midrule
    Fixed & 0.942 & 0.082 & 0.292 & 0.930 & 0.089 & 0.322 & 0.895 & 0.115 & 0.351 \\
    BitFit & 0.945 & 0.080 & 0.290 & 0.931 & 0.089 & 0.321 & 0.902 & 0.108 & 0.347 \\
    Partial-1 & 0.952 & 0.076 & 0.282 & 0.931 & 0.089 & 0.321 & 0.908 & 0.102 & 0.345 \\
    LoRA & 0.949 & 0.076 & 0.284 & 0.934 & 0.087 & 0.317 & 0.912 & 0.098 & 0.344 \\
    Adapter & 0.948 & 0.078 & 0.286 & 0.929 & 0.089 & 0.321 & 0.910 & 0.099 & 0.346 \\
    Adaptformer & 0.947 & 0.078 & 0.287 & 0.931 & 0.089 & 0.321 & 0.912 & 0.098 & 0.345 \\
    Lorand & 0.954 & 0.075 & 0.281 & 0.934 & 0.088 & 0.318 & 0.915 & 0.096 & 0.344 \\
    VFM-Adapter & 0.952 & 0.076 & 0.284 & 0.931 & 0.089 & 0.323 & 0.917 & 0.094 & 0.343 \\ \midrule
    Ours & \textbf{0.963} & \textbf{0.072} & \textbf{0.259} & \textbf{0.941} & \textbf{0.083} & \textbf{0.307} & \textbf{0.918} & \textbf{0.092} & \textbf{0.334} \\
    \bottomrule
    \end{tabular}
\end{table}

\begin{table}[t]
    \centering
    \begin{minipage}{0.48\linewidth}
        \centering
        \caption{Ablation of different intermediate dimensions}
        \label{tab:ablation_hidden_dim}
        \begin{tabular}{c|c|cc}
        \toprule
             \textbf{Dimensions}  & \textbf{Params} & \textbf{mAP} & \textbf{$\text{mAP}_{50}$} \\
             \midrule
             16  & 0.72M & 53.2 & 73.5 \\
             32  & 1.48M & 54.4 & 74.6 \\
             64  & 3.25M & 54.5 & 74.6 \\
             128 & 7.82M & 54.2 & 74.3 \\
        \bottomrule
    \end{tabular}
    \end{minipage}
    \hfill
    \begin{minipage}{0.5\linewidth}
        \centering
        \caption{Performance results of under different Configurations.}
        \label{tab:ablation_conf}
        \begin{tabular}{c|cc}
        \toprule
             \textbf{Configuration} & \textbf{mAP} & \textbf{$\text{mAP}_{50}$} \\
             \midrule
             w/o Scale-Adaptive Fusion   & 52.2 & 73.0 \\
             w/o Semantic Modulation  & 53.6 & 74.1 \\
             All pipeline  & 54.4 & 74.6 \\
        \bottomrule
    \end{tabular}
    \end{minipage}
\end{table}

\paragraph{Universal Adaptability to Diverse Foundation Models.}
SIGMA exhibits consistent superiority across backbones with diverse pre-training paradigms: self-supervised distillation (DINOv2), vision-language contrastive learning (SigLIP2), and promptable geometric modeling (SAM). On SAM, all PEFT methods suffer from inferior performance due to the domain gap between pre-training and downstream tasks. Nevertheless, SIGMA still outperforms VFM-Adapter by 1.1 mAP on detection, 1.22 mIoU on segmentation, and 0.009 RMSE on depth estimation. SIGMA performs well on all three benchmarks, spanning localization (object detection), pixel-level classification (semantic segmentation), and continuous regression (depth estimation). These results demonstrate that the dual-gap problem is a universal challenge in PEFT. By addressing this, SIGMA’s spatial-distributional co-adaptation mechanism overcomes a fundamental limitation of existing PEFT methods, enabling consistent state-of-the-art performance across all tested backbones and tasks.

\paragraph{Bridging the Gap to Full Fine-Tuning.}
SIGMA substantially narrows the performance gap between PEFT methods and full fine-tuning. On DINOv2 detection, the gap between the Fixed baseline and full tuning is 5.1 mAP; SIGMA reduces this to only 1.3 mAP, recovering 74.5\% of the full tuning gain while updating less than 1.7\% of backbone parameters. On ADE20K segmentation with DINOv2, SIGMA achieves 56.42 mIoU compared to 57.10 for full tuning, leaving a margin of only 0.68 mIoU, which is notably smaller than the 1.96 mIoU gap left by VFM-Adapter. On NYUv2 depth estimation with DINOv2, SIGMA reaches a $\delta_1$ accuracy of 0.963 versus 0.968 for full tuning, a difference of merely 0.005. These results indicate that SIGMA effectively addresses the distributional and structural gaps that prevent lightweight adapters from matching full tuning performance.

\subsection{Ablation Studies}

In \cref{tab:ablation_hidden_dim}, we investigate the impact of the intermediate dimension on model performance to determine the optimal trade-off between accuracy and efficiency. Increasing the intermediate dimension from 16 to 32 achieves a substantial improvement, boosting mAP by 1.2 points (53.2 to 54.4) and $\text{mAP}_{50}$ by 1.1 points. However, further increasing the training parameters yields only marginal improvements. Doubling the parameters to 3.25M (dimension 64) results in a negligible mAP gain of 0.1, while further increasing to 128 leads to a slight performance degradation (54.2 mAP). These results indicate that a dimension of 32 effectively captures the necessary semantic information without introducing the redundancy and overfitting risks.

The contribution of each architectural component is validated in \cref{tab:ablation_conf}. The removal of Scale-Adaptive Fusion module triggers the most substantial performance degradation, with mAP dropping from 54.4 to 52.2. This 2.2-point decline highlights the fundamental importance of the filtering mechanism in handling multi-scale features for dense prediction tasks. Furthermore, removing the Semantic Modulation module results in a 0.8 point decrease in mAP (53.6), confirming its necessity for semantic alignment. The superior performance of the complete pipeline demonstrates that these components function synergistically to maximize the adaptation capability of the pre-trained model.

\section{Conclusion}
\label{sec:conclusion}

In this paper, we present SIGMA, a novel parameter-efficient fine-tuning method designed to bridge the gap between large vision foundation models and downstream dense prediction tasks. By integrating Scale-Adaptive Fusion and Semantic Modulation modules, our approach achieves state-of-the-art performance across object detection, semantic segmentation, and depth estimation benchmarks while significantly reducing parameter overhead with only 1.72\% trainable backbone parameters. Extensive experiments and ablation studies on multiple backbones, including DINOv2, SAM, and SigLIP2, demonstrate the method's superior generalization and robustness, establishing SIGMA as a highly efficient and scalable solution for training foundation models in vision tasks.

\section*{Acknowledgments}
This paper uses the MS-COCO dataset (\url{https://cocodataset.org/}, annotations under CC BY 4.0, images under respective Flickr
  Creative Commons licenses), the ADE20K dataset (\url{https://ade20k.csail.mit.edu/}, images for non-commercial research use only, annotations under BSD 3-Clause License), and the NYUv2 dataset (\url{https://cs.nyu.edu/~fergus/datasets/nyu_depth_v2.html}). The authors confirm that the use of all the above datasets in this paper is solely for academic research purposes and has not been used for any commercial activities. We gratefully acknowledge the creators and maintainers of these datasets for making them available to the research community.


%
%
\bibliographystyle{splncs04}
\bibliography{main}
\end{document}